\documentclass[10pt]{article}
\usepackage[preprint]{tmlr}

\usepackage[T1]{fontenc}
\usepackage[utf8]{inputenc}
\usepackage{microtype}
\usepackage{booktabs}
\usepackage{graphicx}
\usepackage{amsmath}
\usepackage{amssymb}
\usepackage{array}
\usepackage[colorlinks=true,linkcolor=blue,citecolor=blue,urlcolor=blue,hypertexnames=false]{hyperref}
\usepackage{url}
\usepackage{xspace}
\usepackage{tikz}
\usetikzlibrary{arrows.meta,positioning}
\usepackage{algorithm}
\usepackage{algpseudocode}

\definecolor{figblue}{HTML}{0072B2}
\definecolor{figorange}{HTML}{E69F00}
\definecolor{figgreen}{HTML}{009E73}

\newcommand{\sigreg}{SIGReg\xspace}
\newcommand{\kv}{K/V\xspace}
\newcommand{\dnll}{$\Delta$NLL\xspace}

\title{Regularize or Localize: When Training-Time KV-Cache \mbox{Geometry} Pays Under Quantization}

\hypersetup{
  pdftitle={Regularize or Localize: When Training-Time KV-Cache Geometry Pays Under Quantization},
  pdfauthor={Libo Sun; Po-Wei Harn; Zewei Zhang; Peixiong He; Xiao Qin},
  pdfsubject={Training-time representation geometry and KV-cache quantization}
}

\author{
\name Libo Sun$^{1}$ \quad
Po-Wei Harn$^{2}$ \quad
Zewei Zhang$^{1}$ \quad
Peixiong He$^{1}$ \quad
Xiao Qin$^{1,*}$ \\
\addr $^{1}$Department of Computer Science and Software Engineering,
Auburn University, Auburn, AL 36830, USA \\
\addr $^{2}$Department of Information Management,
National Central University, Taoyuan 320317, Taiwan \\
\addr \textnormal{Emails: \href{mailto:libo@auburn.edu}{libo@auburn.edu};
\href{mailto:harnpowei@ncu.edu.tw}{harnpowei@ncu.edu.tw};
\href{mailto:zez0001@auburn.edu}{zez0001@auburn.edu}} \\
\addr \textnormal{\phantom{Emails: }\href{mailto:pzh0029@auburn.edu}{pzh0029@auburn.edu};
\href{mailto:xqin@auburn.edu}{xqin@auburn.edu}} \\
\addr \textnormal{$^{*}$Corresponding author: Xiao Qin}
}

\begin{document}
\maketitle

\begin{abstract}
We study whether \sigreg---LeJEPA's anti-collapse objective---can reshape representations during standard autoregressive language-model pretraining, and when the resulting geometry helps \kv-cache quantization. We train 110M-parameter models on 10B FineWeb tokens and report three findings. \textbf{(1)} At $\lambda{=}0.01$, \sigreg reduces hidden-state pairwise-cosine anisotropy by $38\%$ across three paired seeds. Perplexity increases by less than $0.35\%$ in every pair, with no consistent zero-shot loss. \textbf{(2)} This change does not propagate from hidden states to the \kv cache. Applying \sigreg directly to K and V during continued training, however, reduces mean cache anisotropy by $94\%$ across four checkpoints. A matched continuation without the \kv term leaves cache geometry nearly unchanged, and the frozen-trunk retrofits we tested do not reproduce the effect. \textbf{(3)} Under untransformed symmetric group-free quantization, direct \kv regularization is the only training condition that prefers per-channel scaling in all three seeds, and under that same 3-bit per-channel scheme the baseline incurs $4.3$--$7.9\times$ the directly regularized model's \dnll. Under the full simulated KIVI-style configuration (mixed arrangement, zero-points, grouped scales), however, all models reach near-parity, including when storage overhead is approximately matched. In this 110M regime, the training intervention helps when quantizer scales are coarse; the advantage vanishes under the tested combination of token-local grouping, mixed \kv scaling, and zero-points. To our knowledge this is the first training-time \emph{distributional} regularization of standard \kv-cache geometry evaluated against post-hoc cache quantization.
\end{abstract}

\section{Introduction}

Transformer language models trained with cross-entropy develop anisotropic hidden-state geometry: representations concentrate along a few shared directions \citep{ethayarajh2019contextual,gao2019representation,timkey2021rogue}. A separate line of work---joint-embedding predictive architectures (JEPA)---treats representation geometry as a first-class training target, and its recent instantiation LeJEPA \citep{balestriero2025lejepa} introduces \sigreg: a sketched Epps--Pulley normality test \citep{epps1983test} that pushes embeddings toward $\mathcal{N}(0, I)$ along random projections. \sigreg has spread through vision, RL, and time-series models, and has twice been \emph{ablated} on language models with null results---as an auxiliary on a nonstandard architecture's signature stream \citep{akbar2026atma} and on final-layer states under LoRA fine-tuning \citep{sengupta2026representation}. Neither measured geometry; neither targeted the question a systems designer would ask.

We ask that question in three parts. The bridge probes of part~2 use numerical gates fixed before any probe was run; the quantization study of part~3 is exploratory:

\begin{enumerate}
\item \textbf{Does \sigreg shape geometry in standard AR pretraining, and at what cost?} Yes: hidden-state anisotropy falls by $38.0 \pm 5.8\%$ across three paired seeds, at a perplexity cost below $+0.35\%$ in every pair, with capability guards (zero-shot, STS) clean (\S\ref{sec:dose}--\S\ref{sec:guards}).
\item \textbf{Does the geometry extend to the per-layer \kv cache---and can it be retrofitted?} It does not extend on its own, and the frozen-trunk retrofits we tested (new heads on fixed weights) do not recover it. Continued training with \sigreg applied directly to K and V does reduce cache anisotropy---by $94\%$---while a matched continuation without the \kv term does not, isolating the \kv term as the cause (\S\ref{sec:bridge}).
\item \textbf{Does the reduced cache anisotropy buy anything under quantization?} Conditionally. Under untransformed symmetric group-free quantization the directly regularized model is the only training condition that prefers per-channel over per-token scaling, and its 3-bit symmetric per-channel damage is $4.3$--$7.9\times$ smaller than the baseline's; under the full simulated KIVI-style configuration (mixed arrangement, zero-points, grouped scales), all models reach near-parity (\S\ref{sec:quant}).
\end{enumerate}

\textbf{Contributions.} We provide the first anisotropy-targeted characterization of \sigreg in standard LM pretraining, with a paired-seed design that survives a measured $1.8$--$1.9\%$ software-stack drift. We then separate what continued training with the \kv term changes from what frozen-trunk probes and a \kv-term-free continuation do not reproduce, giving a controlled account of where the cache geometry comes from. Third---to our knowledge for the first time---we evaluate a training-time \emph{distributional} regularization of standard \kv-cache geometry against post-hoc cache quantization, with quantizer controls spanning offset absorption to simulated KIVI-style grouped zero-point schemes, and we locate both the payoff and its boundary. Finally, we state explicitly what the pairwise-cosine anisotropy metric does and does not certify, and use that account to interpret the quantization results.

We deliberately do not claim ``first isotropy regularization for LMs'' (that line runs through CosReg, Spectrum Control, SimCTG, I-STAR---\S\ref{sec:related}) nor a deployed compression system (\S\ref{sec:limitations}).

\section{Related Work}
\label{sec:related}

\textbf{JEPA-line objectives for LMs.} LLM-JEPA \citep{huang2025llmjepa} adds a predictive embedding loss over paired views at fine-tuning scale; STP \citep{huang2026stp} predicts semantic-tube triplets on plain token streams. Both retain CE as the implicit anti-collapse mechanism; neither regularizes the embedding \emph{distribution} explicitly. NextLat \citep{teoh2025nextlat} is the nearest methodological neighbor---a next-latent prediction auxiliary in from-scratch pretraining---but shapes no distribution and measures no geometry. NCP \citep{liu2026ncp} discretizes latents into a codebook (CE, not continuous regression). The two \sigreg-on-LM nulls \citep{akbar2026atma,sengupta2026representation} used \sigreg as a generic auxiliary at placements our results predict to be inert or insufficient (final-layer-only; fine-tuning-only; nonstandard stream): our $\lambda{=}0.001$ dose is likewise inert, and our frozen-trunk probes likewise fail---placement and dose carry the effect.

\textbf{Isotropy in LM representations.} The representation-degeneration line regularizes toward isotropy during training: CosReg on output embeddings \citep{gao2019representation}, singular-spectrum control \citep{wang2020spectrum}, token-level contrastive SimCTG \citep{su2022contrastive}. NITP \citep{zhang2026nitp} adds a continuous representation-space auxiliary during pretraining and frames anisotropic drift as the failure mode it prevents---but isotropy there is a byproduct of a predictive objective, geometry is not measured, and \kv is untouched; our target is the distribution itself and the cache. Post-hoc, All-but-the-Top (centering plus top-PC removal) \citep{mu2018allbutthetop} and whitening variants achieve isotropy for free at the embedding level. I-STAR \citep{rudman2024istar} reports that \emph{decreasing} isotropy can help downstream tasks---evidence from encoder fine-tuning, not decoder pretraining; our zero-shot guard addresses the same concern in our regime and finds no sign-consistent cost. Closest to our cache-level observations, \citet{godey2024anisotropy} show anisotropy is inherent to self-attention and that drifted Q/K means are functionally load-bearing---they enable sharp, low-entropy attention---which predicts both that severe K anisotropy is the default and that removing it may cost something (\S\ref{sec:bridge}). We differ from this line in target (per-layer \kv, not just $h$ or output embeddings), mechanism (a distributional normality test, not cosine/spectrum penalties---I-STAR itself observes CosReg acts mainly as a mean shift), and endpoint (cache quantization, not generation diversity).

\textbf{KV-cache compression.} Quantization and low-rank methods treat the trained cache as given: KIVI's asymmetric per-channel-K/per-token-V quantization \citep{liu2024kivi}, KVQuant's pre-RoPE per-channel-K quantization \citep{hooper2024kvquant}, QuaRot's outlier-killing rotations \citep{ashkboos2024quarot}, KVTC's transform coding \citep{kvtc2025}, KV-CoRE's effective-rank benchmarking of cache compressibility \citep{kvcore2026}, and eOptShrinkQ's spectral denoising that ``restores the isotropy scalar quantization assumes'' \citep{eoptshrinkq2026}---all post-hoc. Low-rank attention variants change the cache architecturally (LRKV \citep{lrkv2026}; adaptive-rank compression, STAR-KV \citep{starkv2026}), but none regularize the geometry of a standard attention cache during training. A note on terminology: DMC ``retrofits'' LLMs by continued pretraining \citep{nawrot2024dmc}, which is what our successful probe below does; in this paper ``retrofit'' means frozen-trunk recovery (new heads on fixed weights) or zero-training quantizer changes, with LoRA-scale light adaptation an untested middle ground. Stronger post-hoc normalizations than our controls exist---NSNQuant standardizes \kv vectors channel-wise (mean \emph{and} scale) before quantizing \citep{son2025nsnquant}, and KVarN composes a Hadamard rotation with dual-axis variance normalization for calibration-free 2-bit cache quantization \citep{muller2026kvarn}, whose token-scale-error diagnosis is consonant with our scheme-flip finding---and the K-vs-V asymmetry we observe has independent theoretical grounding: key matrices carry larger norms than values, so keys deserve more bits \citep{hariri2026quantize}.

\textbf{Quantization-friendly training.} A complementary line prevents quantization-hostile geometry during training rather than repairing it afterwards: Outlier-Safe Pre-Training suppresses activation outliers via optimizer and normalization choices \citep{park2025osp}; Quantizable Transformers remove attention-driven outliers architecturally \citep{bondarenko2023quantizable}; gated attention eliminates attention sinks, with quantization cited among the benefits \citep{qiu2025gated}. QAT methods go further and put the quantizer in the training loop, including over the \kv cache \citep{liu2024llmqat,lee2026reqat}. Closest in spirit, KV-CAT trains for token-axis cache \emph{compaction} through \kv-slot sparsification and evaluates shorter continuous caches; it neither targets coordinate-level distributional geometry nor evaluates scalar or low-bit \kv quantization \citep{gelberg2026kvcat}. We share this line's thesis---shape the cache during training---but differ in mechanism and in what is measured: a \emph{distributional} normality regularizer on the standard cache (no optimizer, architecture, sparsification, or in-loop quantizer change), with the geometry itself measured before and after, and the quantizer applied purely post-hoc. Our ``first'' claim is scoped accordingly: first training-time \emph{distributional regularization of \kv-cache geometry} evaluated against post-hoc cache quantization---not first training-time intervention that helps cache compression.

\section{Method}
\label{sec:method}

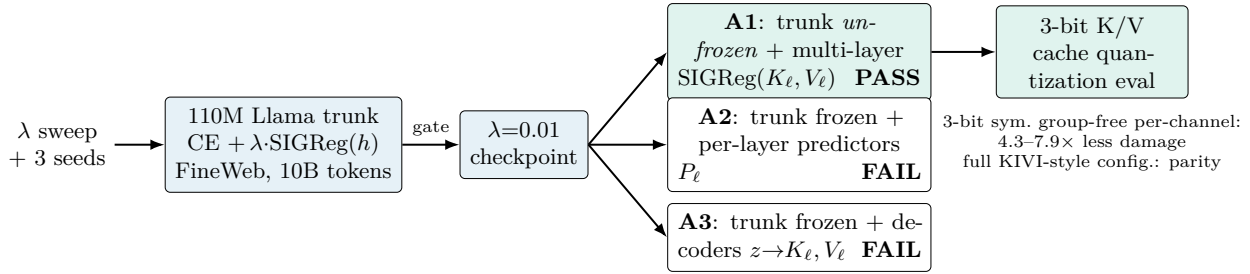
\begin{figure}[t]
\centering
\resizebox{\linewidth}{!}{%
\begin{tikzpicture}[
  font=\small,
  box/.style={draw, rounded corners=2pt, align=center, inner sep=4pt, minimum height=8mm},
  probe/.style={box, text width=34mm},
  arr/.style={-{Latex[length=2mm]}, thick},
]
\node[box, fill=figblue!10] (trunk) {110M Llama trunk\\CE $+\,\lambda\cdot$SIGReg($h$)\\FineWeb, 10B tokens};
\node[left=7mm of trunk, align=center] (data) {$\lambda$ sweep\\$+$ 3 seeds};
\draw[arr] (data) -- (trunk);
\node[box, right=8mm of trunk, fill=figblue!10] (ckpt) {$\lambda{=}0.01$\\checkpoint};
\draw[arr] (trunk) -- (ckpt) node[midway, above, font=\scriptsize] {gate};

\node[probe, right=11mm of ckpt, yshift=13mm, fill=figgreen!12] (a1) {\textbf{A1}: trunk \emph{unfrozen} $+$ multi-layer SIGReg($K_\ell,V_\ell$) \hfill \textbf{PASS}};
\node[probe, right=11mm of ckpt, fill=white] (a2) {\textbf{A2}: trunk frozen $+$ per-layer predictors $P_\ell$ \hfill \textbf{FAIL}};
\node[probe, right=11mm of ckpt, yshift=-13mm, fill=white] (a3) {\textbf{A3}: trunk frozen $+$ decoders $z{\to}K_\ell,V_\ell$ \hfill \textbf{FAIL}};
\draw[arr] (ckpt.east) -- (a1.west);
\draw[arr] (ckpt.east) -- (a2.west);
\draw[arr] (ckpt.east) -- (a3.west);

\node[box, right=9mm of a1, text width=24mm, fill=figgreen!12] (quant) {3-bit \kv cache quantization eval};
\draw[arr] (a1.east) -- (quant.west);
\node[below=1mm of quant, font=\scriptsize, align=center] {3-bit sym.\ group-free per-channel:\\$4.3$--$7.9\times$ less damage\\full KIVI-style config.: parity};
\end{tikzpicture}%
}
\caption{Pipeline. Stage 1 characterizes CE$+\lambda\cdot$\sigreg($h$) pretraining (dose--response $+$ paired seeds). Stage 2 asks whether the \kv cache's geometry can be changed: only the trunk-adapted probe (A1) passes its pre-specified gate; the frozen-trunk probes (A2 at two capacities, A3) fail. Stage 3 measures the payoff of A1's directly regularized cache under post-hoc quantization.}
\label{fig:pipeline}
\end{figure}

Figure~\ref{fig:pipeline} shows the three-stage design; Algorithm~\ref{alg:sigreg} the training-step objective.

\textbf{\sigreg on hidden states (Phase 1).} Following LeJEPA's per-position variant, for hidden states $h \in \mathbb{R}^{B\times T\times d}$ we draw $m$ random unit directions per step, compute the Epps--Pulley sketched-normality statistic of the projected batch distribution at each position (exact estimator and quadrature in Appendix~\ref{app:method}), and add $\lambda\cdot\mathrm{SIGReg}(h)$ to the CE loss. The batch axis is the sample axis of the empirical characteristic function ($B{=}8$); $m{=}1024$, directions resampled every step. The trunk is a 110M Llama-style decoder \citep{touvron2023llama} ($d{=}768$, 12 layers, 12 heads, FFN 2048, vocab 32000, tied head).

\textbf{Bridge probes (pre-specified falsifiers).} From a $\lambda{=}0.01$ checkpoint, three ${\sim}500$M-token probes with numeric gates fixed in the plan before any probe ran:
\begin{itemize}
\item \textbf{A1 (adaptation):} unfreeze the trunk; train CE $+\,\lambda_z\cdot\mathrm{SIGReg}(h)+\lambda_{KV}\cdot\sum_{\ell}[\mathrm{SIGReg}(K_\ell)+\mathrm{SIGReg}(V_\ell)]$ over all layers ($\lambda_{KV}{=}0.01$, $m{=}256$). Gate: mean per-layer \kv pairwise-cosine anisotropy $-30\%$ AND \dnll $\leq 0.15$ nats.
\item \textbf{A2 (frozen prediction):} train per-layer next-position predictors $P_\ell$ on the frozen trunk's \kv. Gate: mean relative residual $< 0.5$. Run at two capacities: linear $d{\to}d$ probes (the plan's original configuration), plus a one-hidden-layer MLP ($2\times$ width, SiLU) capacity check.
\item \textbf{A3 (frozen decoding):} train post-hoc decoders $z \to K_\ell/V_\ell$, where $z$ is the trunk's final hidden state, on the frozen trunk. Gate: attention-score Frobenius error $\leq 0.1$.
\end{itemize}
Probe capacity for A2/A3 was not itself pre-specified---only the thresholds were; we therefore scope every negative as ``at these capacities and this budget'' and add the MLP variant as insurance.

\begin{algorithm}[t]
\caption{One pretraining step of CE $+\,\lambda\cdot$\sigreg($h$) (A1 adds the bracketed lines)}
\label{alg:sigreg}
\begin{algorithmic}[1]
\State sample packed batch $(x, y) \in \mathbb{Z}^{B\times T}$ \Comment{dense, no padding}
\State $h \gets \mathrm{trunk}(x)$; \quad $\mathcal{L} \gets \mathrm{CE}(\mathrm{head}(h),\, y)$ \Comment{[A1: hooks capture $K_\ell, V_\ell\ \forall \ell$]}
\State draw $m$ directions $u_1,\dots,u_m \sim \mathrm{Unif}(\mathbb{S}^{d-1})$ \Comment{fresh every step}
\For{position $t = 1,\dots,T$}
  \State $s_t \gets \frac{1}{m}\sum_{j=1}^{m} \mathrm{EP}\big(\{\,h_{b,t}^{\top} u_j\,\}_{b=1}^{B}\big)$ \Comment{Epps--Pulley statistic over the batch axis}
\EndFor
\State $\mathcal{L} \gets \mathcal{L} + \lambda \cdot \frac{1}{T}\sum_t s_t$ \Comment{[A1: $\lambda \equiv \lambda_z$; $+\,\lambda_{KV}\sum_\ell[\mathrm{SIGReg}(K_\ell) + \mathrm{SIGReg}(V_\ell)]$, same estimator]}
\State AdamW step on $\mathcal{L}$ \Comment{[A1: trunk unfrozen]}
\end{algorithmic}
\end{algorithm}

\textbf{Cache-quantization protocol (exploratory).} We fake-quantize (quantize $\to$ dequantize) each layer's K and V projection outputs at evaluation time, on the same pre-RoPE surface the geometry metrics use---a real design point in deployed systems \citep{hooper2024kvquant}---and report NLL on the standard held-out window as $\Delta$ versus each model's own unquantized reference, plus K-only and V-only diagnostics at 3 bits.

\emph{Scheme definitions.} Group-free per-token quantization uses one scale across channels for each token. Group-free per-channel quantization uses one scale across all batch elements and positions for each channel. With group size $g$, per-token grouping partitions channels into contiguous groups of $g$, while per-channel grouping partitions token positions within each batch element. ``Mixed'' denotes per-channel K with per-token V---the KIVI arrangement \citep{liu2024kivi}. Symmetric grids use absmax scales; asymmetric grids add zero-points, allowing constant offsets to be represented at the same nominal data width (metadata overhead is accounted for separately). The base grid is symmetric and group-free at $b \in \{8,4,3,2\}$ bits; the KIVI-style extension adds the mixed arrangement, grouped scales ($g{=}32$, KIVI's default), and zero-point combinations. Concretely, at $d{=}768$ (12 heads $\times$ 64 dimensions, concatenated), group-free per-token uses one scale per token spanning \emph{all heads}; group-free per-channel uses 768 scales per layer per tensor, each spanning the full batch--sequence extent; $g{=}32$ per-token partitions within 64-dimensional heads; and per-token at $g{=}64$ corresponds to the conventional head-local layout (one scale per head per token). ``Untransformed'' means no basis rotation or channel standardization is applied before scaling. The six group-free configurations (symmetric, zero-point, and mean-subtraction grids at both granularities) form the \emph{original grid}; extensions add mixed, grouped, rotated, and standardized configurations.

\emph{Controls.} Two families guard the comparison. Quantizer-side: zero-point and mean-subtraction controls run on all four training conditions (mean subtraction is dynamic-oracle---statistics from the current full batch, full-precision add-back); a fixed rotation surrogate and dynamic-oracle channel standardization were evaluated on baseline and A1. The training-side control is a token-matched continuation ($\lambda_{KV}{=}0$): the same checkpoint continued for the same 500M tokens with CE $+\,\lambda_z\cdot$\sigreg($h$) only.

\emph{Simulation limits.} This simulates quantizer arrangements, not systems: cache-disabled full-sequence forwards on the pre-RoPE surface (post-RoPE K variants confirm that the group-free findings, the full KIVI-style configuration, and grouped symmetric per-channel all transfer, \S\ref{sec:quant}(i) and (v)), full-precision scale/zero-point metadata, and none of KIVI's block timing or full-precision residual window.

\textbf{Metrics, stated precisely.} Anisotropy is the mean pairwise cosine of sampled representation pairs \citep{ethayarajh2019contextual,mu2018allbutthetop}. It is mean-dominated: removing a shared mean/rogue direction \citep{timkey2021rogue} can drive it to ${\sim}0$ without whitening the covariance. All spectral quantities (top-1 share, condition number, $\sigma_{\min}$) are singular values of the \emph{centered} representation matrix, so they measure covariance shape and cannot be moved by mean removal alone. We co-report the spectral top-1 share throughout (and the condition number at the $h$ level), and treat disagreements between the two metric families as findings (\S\ref{sec:metric}).

\section{Experimental Setup}
\label{sec:setup}

110M Llama-style models pretrained on FineWeb sample-10BT \citep{penedo2024fineweb} for 10B tokens ($610{,}352$ steps $\times\, 8{\times}2048$), lr $3{\cdot}10^{-4}$ cosine (warmup 2000, floor $0.1\times$), AdamW$(0.9, 0.95)$, bf16 autocast, Mistral-7B tokenizer \citep{jiang2023mistral}. Evaluation: 2.01M held-out FineWeb tokens (5000 docs at a 100k-doc offset; the training stream starts after the eval window). The dose--response sweep ran once (seed 42, May 2026 stack); the \{baseline, $\lambda{=}0.01$\} confirmation ran at seeds \{42,43,44\} on a rebuilt venv. The rebuild shifted absolute perplexity by $-1.8\%/{-1.9\%}$ on the two models (near-identical co-movement; seed-42 reproduction check)---all headline statistics are therefore \emph{within-seed, within-stack paired deltas}, and the dose--response sweep is presented as shape evidence only. Zero-shot: lm-eval-harness \citep{gao2023lmeval} (LAMBADA, HellaSwag, PIQA, ARC-easy, WinoGrande) on HF-exported checkpoints whose logits are verified bit-identical to the training-format originals. Compute: one $8{\times}$A100 SLURM node for pretraining (${\sim}33$h per model); all probes, panels, and sweeps on a single RTX~5090.

\section{Results}
\label{sec:results}

\subsection{Dose--response ($n{=}1$, shape)}
\label{sec:dose}

\begin{table}[h]
\centering
\small
\begin{tabular}{lccccc}
\toprule
$\lambda$ & ppl & $\Delta$ppl & aniso($h$) & top1 & cond\# \\
\midrule
0 (baseline) & 21.894 & --- & 0.0242 & 0.0618 & 30.6 \\
0.001 & 21.893 & $-0.0\%$ & 0.0255 & 0.0658 & 33.2 \\
0.01 & 21.934 & $+0.19\%$ & 0.0183 & 0.0556 & 24.3 \\
0.1 & 23.658 & $+8.1\%$ & 0.0107 & 0.0338 & 17.6 \\
1.0 & 22.866 & $+4.4\%$ & 0.0016 & 0.0310 & 32.7 \\
\bottomrule
\end{tabular}
\caption{Dose--response of CE $+\,\lambda\cdot$\sigreg$(h)$ at 10B tokens (seed 42, May stack; $n{=}1$ per arm).}
\label{tab:dose}
\end{table}

\begin{figure}[t]
\centering
\includegraphics{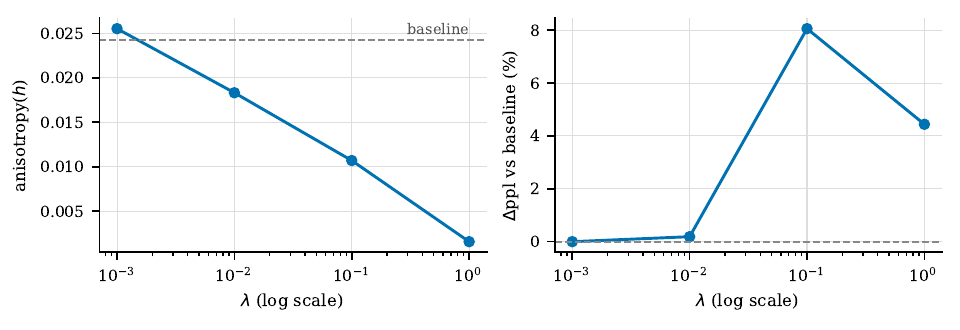}
\caption{Dose--response of \sigreg($h$) at 10B tokens ($n{=}1$ per $\lambda$ setting). Left: anisotropy($h$) falls monotonically over the active doses; $\lambda{=}0.001$ sits on the baseline reference (dashed). Right: in this single-seed sweep the perplexity cost is \emph{not} monotone---$\lambda{=}1.0$ costs less than $\lambda{=}0.1$.}
\label{fig:dose}
\end{figure}

Anisotropy responds monotonically to dose over the active range $\lambda \in \{0.01, 0.1, 1.0\}$ (Figure~\ref{fig:dose}); $\lambda{=}0.001$ is inert (within cross-seed noise, $\pm 0.002$). Perplexity and conditioning are \emph{not} monotone: $\lambda{=}1.0$ reduces the smallest singular value sharply ($\sigma_{\min}$ 62$\to$9), so near-perfect pairwise-cosine isotropy coexists with worse conditioning than baseline---the first of several places the two metric families dissociate. $\lambda{=}0.01$ is the operating point for everything below.

\subsection{Paired confirmation at $\lambda{=}0.01$ ($n{=}3$ seeds)}
\label{sec:paired}

\begin{table}[h]
\centering
\small
\begin{tabular}{lcccc}
\toprule
within-seed $\Delta$ ($\lambda{=}0.01 -$ baseline) & s42 & s43 & s44 & mean $\pm$ sd \\
\midrule
anisotropy($h$) & $-37.4\%$ & $-32.5\%$ & $-44.1\%$ & $\mathbf{-38.0 \pm 5.8\%}$ \\
perplexity & $+0.068\%$ & $+0.179\%$ & $+0.340\%$ & $\mathbf{+0.196 \pm 0.137\%}$ \\
NLL (nats) & $+0.0007$ & $+0.0018$ & $+0.0034$ & $+0.0020 \pm 0.0014$ \\
cond\# & $-7.6$ & $-10.0$ & $-8.4$ & $-8.7 \pm 1.2$ \\
spectral top1 & $+4.4\%$ & $-21.4\%$ & $-23.2\%$ & sign-inconsistent \\
\bottomrule
\end{tabular}
\caption{Within-seed paired deltas, rebuilt stack, $n{=}3$.}
\label{tab:paired}
\end{table}

The anisotropy effect is sign-consistent and ${\sim}6.5\times$ the pair-to-pair spread; the perplexity cost is positive in every pair and below $+0.35\%$ (largest pair $+0.340\%$). We do not report seed-level inferential statistics at $n{=}3$; the claim is the descriptive bound and its sign-consistency (within-seed quantization margins later carry eval-window bootstrap intervals, \S\ref{sec:quant}(iii)). The $\lambda$-arm's across-seed anisotropy spread ($\pm 0.0007$) is $3\times$ tighter than baseline's ($\pm 0.0020$): the regularizer pins the metric it targets. The spectral top-1 share moves inconsistently---at this dose \sigreg acts primarily on shared directions, with only weak, seed-inconsistent pressure on the top of the spectrum, though the conditioning improvement \emph{is} sign-consistent ($-8.7 \pm 1.2$)---the estimator intuition in \S\ref{sec:metric} is consistent with exactly this diluted-but-nonzero spectral pressure.

\subsection{Capability guards}
\label{sec:guards}

\textbf{Zero-shot (5 tasks $\times$ 3 seeds).} No task shows a sign-consistent within-seed effect; per-task mean deltas (percentage points): LAMBADA $+0.81$, HellaSwag $+0.03$, PIQA $+0.53$, ARC-e $-0.01$, WinoGrande $-1.45$---each within the standard error of a between-model comparison ($\sqrt{2}\,\times$ single-run task SE; the largest single-pair excursion is WinoGrande $-4.2$pp at s43, within cross-seed task noise). The regularizer changes geometry, not measured zero-shot capability.

\textbf{STS-B and the All-but-the-Top comparator.} Raw pooled-embedding STS-B Spearman improves within-seed in all three pairs ($0.559{\to}0.594$, $0.542{\to}0.577$, $0.538{\to}0.564$). But All-but-the-Top post-processing (centering plus top-7-PC removal) lifts \emph{every} model, baseline included, to $0.688$--$0.697$. At the hidden-state level, this post-hoc centering-plus-PC-removal subsumes the trained regularizer's sentence-embedding benefit---we state this plainly, because it localizes the interesting question at the \kv level, where the same post-hoc logic fails (\S\ref{sec:bridge}).

\subsection{Bridge characterization: adaptation vs.\ retrofit}
\label{sec:bridge}

\textbf{A1 (adaptation) passes on four of four pretraining checkpoints.} A1 was evaluated from four distinct starting checkpoints (the dose--response $\lambda{=}0.01$ checkpoint and the three seed-confirmation checkpoints); two runs were repeated in the same configuration to persist the adapted weights, giving six runs total (all probes ran on the local stack, on checkpoints from both stacks). Across these runs, mean per-layer \kv anisotropy fell by a $0.936$--$0.943$ fraction (gate $\geq 0.30$) at a $+0.095$--$0.101$-nat NLL cost (guard $\leq 0.15$). At tensor level, 143 of 144 per-run layer-tensor checks clear the 30\% bar; the single exception is V layer~0 at $27.0\%$ in the s43 run. V layer~0 is the weakest tensor in every run ($40.7$--$44.6\%$ in the other five; dose-checkpoint run minimum $43.9\%$ with all 24 tensors clear). Before adaptation the trunk's K is severely anisotropic (per-layer pairwise-cosine ${\sim}0.76$ mean; overall \kv 0.42) despite the $h$-level regularizer---$h$-level anisotropy reduction does not propagate to the cache by itself, consistent with K anisotropy being self-attention's default operating point \citep{godey2024anisotropy}.

\textbf{The token-matched continuation control isolates the cause.} The same checkpoints continued for the same 500M tokens \emph{without} the \kv term ($\lambda_{KV}{=}0$; one run per seed-confirmation checkpoint, $n{=}3$) reduce \kv anisotropy by only a $0.04$--$0.05$ fraction in every run. The matched hidden-\sigreg continuation does not produce the geometry change; adding the multi-layer \kv term does.

\textbf{A2 (frozen prediction) fails at two capacities} (single runs on the dose--response checkpoint). Mean relative residual $0.618$ (linear) and $0.593$ (MLP $2\times$, SiLU) against a $<0.5$ gate. The decomposition is the finding: K mean residual $0.37$ (linear) / $0.35$ (MLP)---$0.28$--$0.35$ across layers 1--11 with layer~0 an outlier at $0.86$--$0.88$---versus V mean $0.87$ (linear) / $0.84$ (MLP). K alone would clear the gate; V stays ${\sim}70\%$ above it at both capacities, and doubling probe capacity moves it by only $0.03$. Next-position K is largely predictable from a frozen trunk; V is not.

\textbf{A3 (frozen decoding) fails decisively} (single run, same checkpoint). Attention-score Frobenius error $0.861$ vs.\ a $0.1$ gate ($8.6\times$ the threshold; layers 3--11 near-total error). Post-hoc reconstruction of usable K from the trunk's output embedding is not available at this budget and capacity.

Reading: the \kv geometry of a CE($+$\sigreg-on-$h$) trunk can be changed by \emph{continued training with trunk adaptation} at a bounded, replicated quality cost, but was not recovered by the frozen-trunk probes we tested (linear and MLP-$2\times$, one checkpoint, matched 500M-token budget). The corroborating external evidence is the pair of published \sigreg-on-LM nulls at other placements \citep{akbar2026atma,sengupta2026representation}: where the regularizer acts determines whether it does anything.

\subsection{The payoff: cache quantization ($n{=}3$, exploratory)}
\label{sec:quant}

\begin{table}[h]
\centering
\small
\begin{tabular}{lcccc}
\toprule
bits & baseline & \sigreg($h$) & A1-adapted & baseline/A1 per-seed \\
\midrule
8 & $+0.001$ & $+0.000$ & $+0.000$ & --- \\
4 & $+0.219$ & $+0.204$ & $\mathbf{+0.070}$ & $2.4$--$4.2\times$ \\
3 & $+1.659$ & $+1.289$ & $\mathbf{+0.268}$ & $\mathbf{4.3}$--$\mathbf{7.9\times}$ \\
2 & $+6.92$ & $+3.94$ & $+4.02$ & (all unusable) \\
\bottomrule
\end{tabular}
\caption{\dnll vs.\ each model's own fp32 reference under symmetric group-free per-channel \kv fake quantization; mean over 3 seeds.}
\label{tab:quant}
\end{table}

\begin{figure}[t]
\centering
\includegraphics{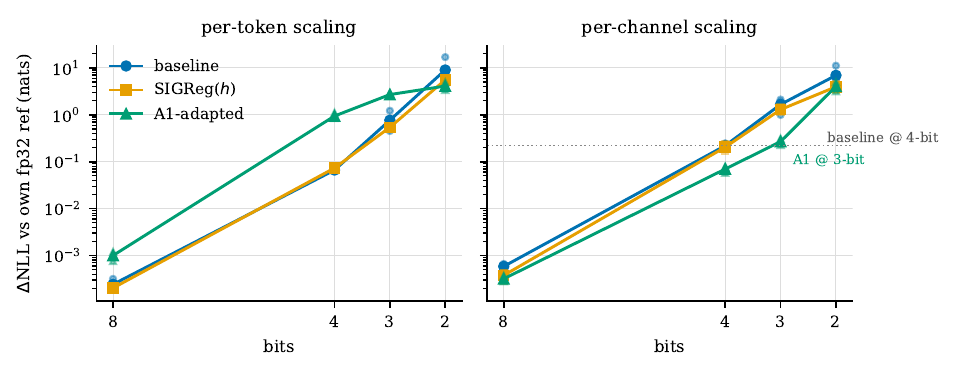}
\caption{The scheme flip (untransformed symmetric group-free scales). \dnll (log scale) vs.\ quantization bit-width for both scale granularities; lines are 3-seed means, faint points individual seeds. Per-token (left): the baseline's preferred scheme, on which A1 is \emph{worst}. Per-channel (right): A1 dominates at 3--4 bits (4.3--7.9$\times$ less degradation at 3 bits). The full simulated KIVI-style configuration compresses these differences to near-parity (Table~\ref{tab:kivi}).}
\label{fig:quant}
\end{figure}

\begin{table}[t]
\centering
\small
\begin{tabular}{lcccc}
\toprule
3-bit scheme & baseline & \sigreg($h$) & A1-adapted & continuation ($\lambda_{KV}{=}0$) \\
\midrule
symmetric, per-token & $+0.765$ & $+0.540$ & $+2.693$ & $+0.341$ \\
symmetric, per-channel & $+1.659$ & $+1.289$ & $\mathbf{+0.268}$ & $+0.591$ \\
asymmetric, per-token & $+0.736$ & $+1.182$ & $+2.536$ & $+0.479$ \\
asymmetric, per-channel & $+1.810$ & $+1.744$ & $+1.533$ & $+0.607$ \\
mean-sub $+$ sym., per-token & $+2.299$ & $+2.949$ & $+4.434$ & $+0.929$ \\
mean-sub $+$ sym., per-channel & $+2.348$ & $+2.672$ & $+0.748$ & $+1.274$ \\
\bottomrule
\end{tabular}
\caption{Controls at 3 bits: \dnll vs.\ own fp32 reference under the six displayed all-model configurations (symmetric, zero-point, and mean-subtraction grids at both granularities; 3-seed means in every column). No mean-absorbing retrofit rescues the baseline's per-channel column or flips its preferred scheme; only A1 flips.}
\label{tab:controls}
\end{table}

Five observations (Figure~\ref{fig:quant}, Tables~\ref{tab:quant}--\ref{tab:kivi}).

\textbf{(i) A1 uniquely changes the preferred scaling axis under untransformed symmetric group-free quantization.} We call this the \emph{scheme flip}: a change in preferred scaling from per-token to per-channel. The baseline prefers per-token scaling (3-bit: $+0.76$ vs.\ $+1.66$ per-channel)---token-norm variation across positions inflates shared per-channel scales \citep{xiao2023streamingllm}; our concentration proxy (\S\ref{sec:limitations}) is mild at this scale, but these scales pool the full batch--sequence extent---and so do the $h$-only model and the continuation control, in every seed; A1 is the only training condition that prefers per-channel in all three seeds. Among the controls we ran, none makes the baseline share this preference. Zero-point grids barely move the baseline ($+0.74$ per-token) and do not repair its per-channel configuration ($+1.81$); dynamic-oracle mean subtraction makes every training condition worse (baseline per-token $+2.30$)---an empirical regularity of our grids: mean-removed geometry suits per-token absmax scaling poorly whether the mean is removed by arithmetic or by training (\S\ref{sec:metric} supplies channel-level correlates for the trained case). A fixed rotation surrogate (random-orthogonal basis---dense, not the Hadamard-structured, head-aware, post-RoPE transform deployed QuaRot uses; evaluated on baseline and A1) substantially improves the baseline's zero-point per-token configuration ($+0.74 \to +0.18$ mean at 3 bits, spanning $+0.10$--$+0.28$ across seeds) while degrading per-channel scaling for both (channel identity does not survive rotation); dynamic-oracle channel standardization (mean and scale recomputed on the current full batch, future tokens included---a favorable, non-deployable diagnostic) backfires even so (baseline symmetric per-token $+0.77 \to +1.87$). These transforms change the geometry the scales see, and preferences with them: rotated or standardized, the baseline still prefers per-token, and A1's own best configuration under the extended grid becomes standardize$+$rotate per-token ($+0.13$--$+0.17$, sharing the high-level rotation-plus-normalization ingredients used by KVarN \citep{muller2026kvarn}). The single-target diagnostics localize A1's per-token damage to V (K-only $+1.17$ / V-only $+2.36$ vs.\ baseline $+0.38$/$+0.10$): V is mean-reduced but spectrally \emph{more} concentrated---poorly suited to per-token scales, well suited to per-channel. The flip is specifically a coarse-extent phenomenon: under the conventional head-local layout (per-token at $g{=}64$, one scale per head per token), every model---A1 included---prefers per-head per-token over group-free per-channel, and A1's advantage persists there as magnitude rather than axis (it is best on every per-head configuration: 3-bit symmetric $+0.127$ vs.\ the baseline's $+0.229$; zero-point $+0.055$ vs.\ $+0.101$; $1.8\times$ in the means). The quantization locus transfers: quantizing K \emph{after} rotary embedding (the surface deployed caches store; V has no RoPE) reproduces the group-free structure nearly quantitatively---3-bit symmetric per-channel baseline $+1.807$ vs.\ A1 $+0.331$ ($5.5\times$; pre-RoPE $6.2\times$), per-token ordering intact for both models---so the pre-RoPE simulation is not load-bearing for the group-free findings.

\textbf{(ii) The symmetric per-channel advantage is robust across seeds.} Scheme-matched at 3 bits, A1's degradation is $4.3$--$7.9\times$ smaller than the baseline's in every seed ($2.4$--$4.2\times$ at 4 bits; Table~\ref{tab:quant}). At this scheme A1 at 3 bits degrades about as much as the baseline at 4 bits ($+0.268$ vs.\ $+0.219$ in the mean)---informally, one extra bit at iso-degradation. Its damage magnitude is also stable: $+0.23$--$0.31$ across seeds against the baseline's $+0.99$--$2.11$ (and $+4.1$--$11.1$ at 2 bits). The baseline's ranking of configurations is itself seed-dependent: its best configuration changes identity with seed on the original grid (with rotation admitted it stabilizes on rotated per-token), and, separately, its asymmetric (zero-point) mixed-arrangement value spans $6\times$. The $h$-only \sigreg model sits between baseline and A1 at the usable 3--4-bit symmetric per-channel points in the three-seed means---per-seed the ordering crosses twice (s43 at 3 bits: $1.17$ vs.\ the baseline's $0.99$; s44 at 4 bits: $0.24$ vs.\ $0.22$)---a dose narrative from $h$-regularization to full \kv adaptation, scheme-conditional throughout.

\textbf{(iii) Best-configuration rankings depend on bit width and the quantizer grid.} Define the \emph{margin} as the baseline's best-configuration NLL minus A1's best-configuration NLL over a stated grid, in absolute nats: positive favors A1, which carries its $+0.10$-nat unquantized handicap into this accounting. On the original group-free grid at 3 bits, A1 wins in every seed (margins $+0.011$/$+0.125$/$+0.261$); a paired packed-batch bootstrap that reselects each model's best configuration inside every replicate places the 95\% interval above zero even for the smallest margin (s42 $[+0.004, +0.018]$, bootstrap win frequency $0.999$; Appendix~\ref{app:method}). On the same six-configuration group-free grid at 4 bits the ranking reverses: margins $-0.101$/$-0.121$/$-0.103$, all intervals excluding zero---4-bit damage is too small to repay the handicap---so the one-extra-bit reading from (ii) holds per-channel scheme-matched but not best-configuration-vs-best-configuration. Extending the 3-bit grid changes the ranking again: the asymmetric group-free KIVI arrangement gives the baseline $+0.149$/$+0.319$/$+0.925$ \dnll across seeds---a $6\times$ spread on a single configuration---and with the mixed arrangement and selected competitive rotation/standardization configurations included the margins become $-0.139$ ($[-0.143,-0.135]$) at s42, $-0.104$ ($[-0.108,-0.101]$) at s43, and $+0.050$ ($[+0.045,+0.056]$) at s44. Each seed's winner is resolved at held-out-window resolution; the ranking itself is seed- and grid-dependent. What is robust is the preference (i), the scheme-matched advantage (ii), and where the contest dissolves (v).

\textbf{(iv) The \kv-term-free continuation already shows part of the raw robustness---not the preference.} The $\lambda_{KV}{=}0$ continuation ($n{=}3$, one run per seed checkpoint) retains the hidden-state term (CE $+\,\lambda_z\cdot$\sigreg($h$)); its cache anisotropy-reduction is $0.04$--$0.05$ in every run, against A1's $0.94$ (same convention: fraction of anisotropy removed), and it costs $+0.073$--$+0.076$ nats of its own---so roughly three quarters of A1's $+0.10$-nat full-precision penalty is already present in the matched \kv-term-free hidden-\sigreg continuation. This design does not distinguish hidden-state regularization, additional training, and the learning-rate restart as the source. The continuation nevertheless gains substantial quantization robustness (original-grid best configurations $+0.27$--$+0.35$ across seeds, vs.\ the baseline's $+0.42$--$+0.63$ and A1's $+0.23$--$+0.31$---a near-tie against A1 once handicaps are counted), so raw low-bit tolerance does not require the \kv term. What it does not gain, in any seed: its preferred scaling is still per-token, and its same-seed symmetric per-channel configuration is $2.1$--$2.4\times$ worse than A1's ($+0.54$--$+0.66$ vs.\ $+0.23$--$+0.31$). The \kv term's specific contribution is the geometry (\S\ref{sec:bridge}), the scheme flip, and the per-channel operating point---not the bulk of the raw best-configuration robustness.

\begin{table}[t]
\centering
\small
\begin{tabular}{lcccc}
\toprule
3-bit KIVI-style scheme (simulated) & baseline & \sigreg($h$) & A1-adapted & continuation \\
\midrule
mixed (pc-K/pt-V), symmetric, group-free & $+0.793$ & $+0.641$ & $+2.439$ & $+0.332$ \\
symmetric, per-token, per-head ($g{=}64$) & $+0.229$ & $+0.212$ & $+0.127$ & $+0.149$ \\
asymmetric, per-token, per-head ($g{=}64$) & $+0.101$ & $+0.097$ & $+0.055$ & $+0.072$ \\
symmetric, per-token, $g{=}32$ & $+0.129$ & $+0.125$ & $+0.084$ & $+0.090$ \\
symmetric, per-channel, $g{=}32$ & $+1.045$ & $+2.045$ & $\mathbf{+0.105}$ & $+0.249$ \\
mixed (pc-K/pt-V), symmetric, $g{=}32$ & $+0.161$ & $+0.113$ & $+0.036$ & $+0.077$ \\
asymmetric, per-token, $g{=}32$ & $+0.057$ & $+0.055$ & $+0.036$ & $+0.042$ \\
KIVI-style: mixed, asym., $g{=}32$ & $+0.016$ & $+0.015$ & $+0.015$ & $+0.012$ \\
\bottomrule
\end{tabular}
\caption{KIVI-style quantizers at 3 bits nominal (\dnll vs.\ own fp32 reference; 3-seed means in every column). Grouping alone (symmetric rows) preserves A1's per-channel advantage ($9.9\times$ in the mean) and a modest per-token edge; the full KIVI-style configuration brings every model to near-parity. Grouped zero-point scales add an analytical ${\sim}1$ bit/value of metadata ($2\times 16$-bit parameters per 32 elements; kept full-precision and free to all models in this simulation).}
\label{tab:kivi}
\end{table}

\textbf{(v) The full KIVI-style configuration removes the advantage at nominal bit width.} Table~\ref{tab:kivi}: quantizer-side structure absorbs the geometry advantage in stages---a quantizer-side dose--response mirroring the training-side one. Per-head scales (per-token, $g{=}64$) preserve a $1.8\times$ A1 edge; grouped scales at $g{=}32$ (symmetric) still favor A1 by $9.9\times$ per-channel ($+0.105$ vs.\ $+1.045$), $1.5\times$ per-token ($+0.084$ vs.\ $+0.129$), and $4.5\times$ in the mixed arrangement ($+0.036$ vs.\ $+0.161$). Zero-points complete the collapse: asymmetric grouped per-token narrows to $+0.036$ vs.\ $+0.057$. A batch-isolation control locates the driver: per-channel scales at \emph{full-sequence} extent within each batch element ($g{=}T$, no cross-batch pooling, no locality change) leave the group-free picture intact---baseline $+1.34$ vs.\ A1 $+0.22$, $6.2\times$, matching the batch-pooled $6.2\times$---so what removes the advantage is scale locality \emph{along the token axis}, together with the mixed arrangement and zero-points, not cross-batch scale sharing. The full KIVI-style configuration (mixed arrangement $+$ zero-points $+$ $g{=}32$) brings every model to $+0.011$--$+0.016$ per seed at 3 bits nominal ($+0.003$ at 4)---near-parity (used descriptively throughout: across-model spreads of ${\sim}0.005$ nats at 3 bits and ${\sim}0.03$ at nominal-2-bit, against group-free spreads above 1 nat; no formal equivalence test is implied), at which point A1's $+0.10$-nat base handicap dominates end-to-end NLL and the training-time intervention is a net loss in this regime at this scale. The boundary holds at the deployed K locus: applying the full KIVI-style configuration with post-RoPE K lands every model at $+0.017$--$+0.021$ at 3 bits nominal ($+0.004$ at 4), and post-RoPE grouped symmetric per-channel preserves A1's advantage ($+0.11$ vs.\ the baseline's $+1.06$, $9.7\times$)---both sides of the story transfer. The storage-adjusted comparison cuts both ways: grouped zero-point scales are not free at deployment (an analytical ${\sim}1$ extra bit per value, making 3-bit $g{=}32$ storage comparable to group-free 4-bit), and our grids keep that metadata full-precision and uncharged for every model. The comparison still lands the same way: \emph{nominal-2-bit} KIVI-style configurations (analytically ${\approx}3.0$ bits/value) sit at model means $+0.075$--$+0.105$---near-parity again, and at or below every model's best group-free 3-bit configuration (per-seed there is one exception: the baseline's rotated best at s42, $+0.099$, edges its own 2-bit grouped $+0.111$). The training-time advantage is a \emph{coarse-scale-regime} property: real where quantizer scales span long token extents; it vanishes under the tested combination of token-local grouping, mixed \kv scaling, and zero-points.

\section{What the anisotropy metric certifies---and what it doesn't}
\label{sec:metric}

\begin{figure}[t]
\centering
\includegraphics{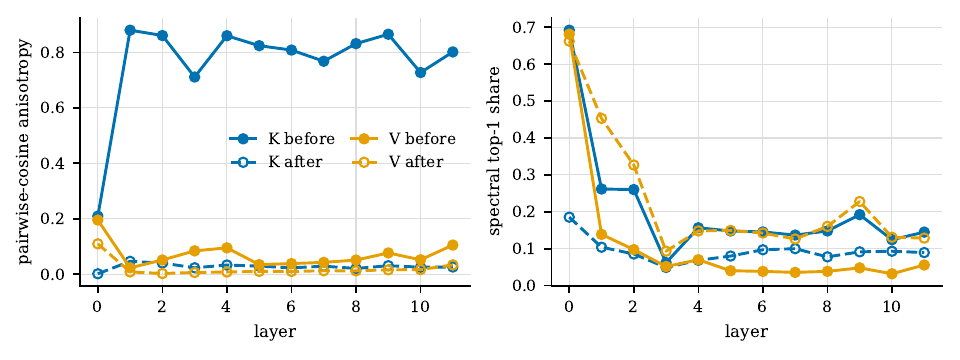}
\caption{Per-layer \kv geometry before (solid) and after (dashed, open markers) the A1 probe, primary run. Left: pairwise-cosine anisotropy collapses for every layer's K and V. Right: the spectral top-1 share \emph{dissociates}---K flattens while V concentrates, especially in early layers.}
\label{fig:perlayer}
\end{figure}

The geometry gates in this paper (Phase 1 and A1) use mean-pairwise-cosine anisotropy, which is mean-dominated: it indicates reduced average directional alignment---a quantity a shared mean or rogue direction can dominate---and does not certify removal of any particular direction, nor covariance whitening. Our results show both faces. At the $h$ level ($\lambda{=}0.01$), top-1 spectral share is sign-inconsistent across seeds while pairwise cosine drops 38\%---consistent with a predominantly mean-level change, since our spectral quantities are centered (\S\ref{sec:method}) and mean removal cannot move a centered spectrum. After A1 (primary run, Figure~\ref{fig:perlayer}), the \emph{aggregate} \kv top-1 share is flat ($0.158{\to}0.161$), but per-layer it decomposes into K $-55\%$ ($0.206{\to}0.094$, genuine spectral flattening) and V $+107\%$ ($0.111{\to}0.229$; layer 1: $0.139{\to}0.453$---spectral \emph{concentration}). We could have buried this in an appendix; instead it is our account of \S\ref{sec:quant}'s scheme flip, and the channel-level statistics supply strong, consistent \emph{correlates} in the basis quantization actually uses (correlates, not a causal identification): after A1, V's per-channel scale dispersion (pooled median of per-layer max/median channel stds) rises to $8.5$ against $1.9$--$2.2$ for every other model---consistent with greater sensitivity of per-token absmax scaling to a small number of high-variance channels---while K's heaviest channel tail collapses (pooled median of per-layer max excess kurtosis $50 \to 12$). It also sharpens an open question: whether the increased V spectral concentration helps or harms low-rank cache compression is untested, and is distinct from the quantization robustness studied here.

The dissociation is, moreover, \emph{consistent with} the estimator's structure---an intuition about sensitivity, not a statement about training gradients. The per-position Epps--Pulley statistic is computed over the batch axis---$B{=}8$ samples---where a normality test may have limited sensitivity to distributional \emph{shape}; it is sensitive mainly to the projected first and second moments. And, heuristically, for a random unit direction $u$, $\mathbb{E}_u[\langle u, v_1\rangle^2] = 1/d$, so a rank-one covariance spike along $v_1$ contributes on average $1/d$ of its energy to $u^\top \Sigma u$ ($d{=}768$), suggesting weak, not absent, pressure on any individual spectral direction. That is what the data show: a strong reduction in mean/rogue-direction alignment without whitening, alongside small but sign-consistent second-moment movement (the $h$-level condition number improves in all three pairs, Table~\ref{tab:paired}). What this intuition does not predict is the \emph{direction} of the large per-layer spectral moves after A1---K's $-55\%$ flattening, V's $+107\%$ concentration; those are facts the account accommodates, not ones it forecasts. All such attributions are intervention-level: adding the \kv objective changes geometry and quantization outcomes jointly; mediation by any individual geometric statistic---pairwise-cosine anisotropy included---is not established. Claims in this paper are phrased accordingly.

\section{Limitations}
\label{sec:limitations}

\textbf{Scale and generalization.} All claims are scoped to one regime: 110M parameters, one corpus (FineWeb 10BT), one tokenizer and architecture. Modern GQA/MQA cache layouts, whose per-KV-head geometry and outlier structure differ, are untested. Near-parity under the full KIVI-style configuration is a result at this scale---and our channel statistics suggest why scale matters: our token-norm concentration proxy (median max/median per-token \kv norm $1.2$--$1.7$; no value above $3.6$) shows no severe concentration, in contrast to the extreme outlier structure reported at larger scale. The proxy measures norms, not attention mass, so it is not a direct attention-sink measurement \citep{xiao2023streamingllm}. The same caveat bounds our positive results symmetrically.

\textbf{Probe evidence.} The A2/A3 negatives are single runs on a single checkpoint, bounded by probe capacity (linear, MLP-$2\times$) and a 500M-token budget; the pre-specification covers thresholds, not capacities, and LoRA-scale light adaptation between ``frozen'' and ``fully unfrozen'' is an untested middle ground. Two of the six A1 runs are same-configuration re-runs rather than independent replicates, and the $\lambda_{KV}{=}0$ continuation control has one run per seed checkpoint ($n{=}3$). A1 costs $+0.10$ nats of base NLL, ${\sim}50\times$ the Phase-1 regularizer's cost; whether a gentler adaptation schedule shrinks it is open---if drifted K/Q means are load-bearing for sharp attention \citep{godey2024anisotropy}, part of the cost may be the price of re-routing attention around the removed drift, a hypothesis we do not test.

\textbf{Simulated deployment.} Quantization is simulated (fake quantization in cache-disabled full-sequence forwards, NLL-only) and was not pre-specified in the plan---no deployed cache and no wall-clock or memory measurements. Evaluation is prefill-style throughout; autoregressive error accumulation under decoding, which motivates pseudo-decode evaluation in recent work \citep{muller2026kvarn}, is untested. The post-RoPE K locus is verified for the group-free grid (\S\ref{sec:quant}(i)), the full KIVI-style configuration, and grouped symmetric per-channel (\S\ref{sec:quant}(v)); the remaining grouped cells are pre-RoPE only. The simulated KIVI-style grouped grid charges no model for scale/zero-point metadata storage; the iso-storage configurations (\S\ref{sec:quant}(v)) address metadata overhead under the stated analytical accounting, not with quantized metadata. Our rotation control is a dense random-orthogonal \emph{surrogate} for the QuaRot class \citep{ashkboos2024quarot}---not Hadamard-structured, head-aware, fused, or post-RoPE as deployed---and involves no statistics; our standardization control (alone or composed with rotation) uses dynamic oracle statistics computed on the full current batch (future tokens included; degenerate at batch-of-one decoding), covering NSNQuant's normalization idea but not its calibration-free causal pipeline \citep{son2025nsnquant}. The oracle treatment strengthens the standardization-backfire result but makes those configurations favorable, non-deployable diagnostics rather than deployable retrofits; both controls also ran only on baseline and A1.

\textbf{Statistical scope.} $n{=}3$ supports descriptive bounds across seeds, not seed-level inference; \emph{within}-seed best-vs-best margins carry paired-bootstrap intervals at held-out-window resolution (\S\ref{sec:quant}(iii), Appendix~\ref{app:method}). Part of A1's raw low-bit tolerance is already present in the \kv-term-free hidden-\sigreg continuation (\S\ref{sec:quant}(iv)); the geometry, the scheme flip, and the per-channel operating point are the \kv-term-specific residue. The two software stacks are never pooled; the $-1.8\%/{-1.9\%}$ drift is disclosed and quarantined by the paired design.

\section{Conclusion}

\sigreg provides a low-NLL-cost way to reduce shared-direction anisotropy during standard autoregressive pretraining. This change does not automatically propagate to the \kv cache. Direct \kv regularization during continued training reduces cache anisotropy by $94\%$, whereas a matched continuation without the \kv term and the frozen-trunk probes we tested do not reproduce the effect.

The quantization consequence depends on scale granularity. Under untransformed symmetric group-free scaling, the \kv-regularized model is the only training condition that prefers per-channel scaling, and the baseline incurs $4.3$--$7.9\times$ as much \dnll under 3-bit symmetric per-channel quantization. This advantage disappears under the full simulated KIVI-style configuration (mixed arrangement, zero-points, and $g{=}32$ grouping), including when storage overhead is approximately matched under our analytical metadata accounting. Training-time regularization therefore controls cache geometry; in the tested grid, its quantization advantage appears with long-span scales and disappears under that full token-local grouped zero-point configuration.

\bibliographystyle{tmlr}
\bibliography{refs}

\appendix

\vspace{2ex}
\noindent\rule{\linewidth}{0.6pt}
\begin{center}
{\Large\bf Appendix}
\end{center}
\vspace{-1ex}

\section{Numbers provenance}
\label{app:provenance}

Every quantitative claim traces to an artifact in the code release (paths repo-relative). Curated copies of every artifact below ship under version control in \texttt{paper/artifacts/} with a manifest recording sha256 hashes (the collection-time git revision is redacted from the bundle for double-blind review and logged externally; \texttt{scripts/collect\_paper\_artifacts.py}); table bodies regenerate from them via \texttt{scripts/paper\_tables.py}.

\begin{table}[h]
\centering
\footnotesize
\begin{tabular}{l >{\raggedright\arraybackslash}p{0.48\linewidth}}
\toprule
claim & artifact \\
\midrule
\S\ref{sec:dose} dose--response & \texttt{outputs/phase1\_exitgate/sweep\_metrics.json} \\
\S\ref{sec:paired} paired table; stack drift & \texttt{outputs/phase1\_seedconfirm\_aggregate.json} \\
\S\ref{sec:guards} zero-shot & \texttt{outputs/zeroshot/*/} \\
\S\ref{sec:guards} STS $+$ All-but-the-Top & \texttt{outputs/sts\_probe/results*.json} \\
\S\ref{sec:bridge} A1 $\times 6$ & \texttt{outputs/phase1\_5/a1/}, \texttt{outputs/phase1\_5\_replicates/*/a1/} \\
\S\ref{sec:bridge} A2 linear / MLP & \texttt{outputs/phase1\_5/a2/}, \texttt{outputs/phase1\_5\_replicates/a2\_mlp\_h2/} \\
\S\ref{sec:bridge} A3 & \texttt{outputs/phase1\_5/a3/} \\
\S\ref{sec:quant} quant grid $+$ diagnostics & \texttt{outputs/kv\_quant/results\_s*.json} \\
\S\ref{sec:quant} mean-absorbing controls & \texttt{outputs/kv\_quant/controls\_s*.json} \\
\S\ref{sec:quant} KIVI-style grouped grid (Table~\ref{tab:kivi}) & \texttt{outputs/kv\_quant/results\_kivi\_*.json} \\
\S\ref{sec:quant} rotation/standardization retrofits & \texttt{outputs/kv\_quant/results\_retrofit\_main.json} \\
\S\ref{sec:quant}(i) post-RoPE K locus & \texttt{outputs/kv\_quant/results\_postrope\_main.json} \\
\S\ref{sec:quant}(i) per-head ($g{=}64$) control & \texttt{outputs/kv\_quant/results\_perhead\_all.json} \\
\S\ref{sec:quant}(v) post-RoPE KIVI-style & \texttt{outputs/kv\_quant/results\_postrope\_kivi\_all.json} \\
\S\ref{sec:quant} $\lambda_{KV}{=}0$ continuation $\times 3$ & \texttt{outputs/phase1\_5\_replicates/s4*\_ce\_continue/} \newline \texttt{outputs/kv\_quant/*ce\_continue*.json} \\
\S\ref{sec:quant}(v) iso-storage 2-bit configurations & \texttt{outputs/kv\_quant/results\_kivi2bit\_all.json} \\
\S\ref{sec:quant}(v) batch-isolation ($g{=}T$) control & \texttt{outputs/kv\_quant/results\_gcontrol\_all.json} \\
\S\ref{sec:quant}(iii) bootstrap margins & \texttt{outputs/kv\_quant/bootstrap\_\{records,summary\}.json} \\
\S\ref{sec:metric} per-layer \kv top-1 & \texttt{outputs/phase1\_5/a1/probe\_result.json} \\
\S\ref{sec:metric} channel statistics & \texttt{outputs/channel\_stats/results\_all.json} \\
export logits-equivalence & \texttt{outputs/hf\_export/*/export\_manifest.json} \\
\bottomrule
\end{tabular}
\caption{Claim-to-artifact map.}
\label{tab:provenance}
\end{table}

\section{Estimator and probe definitions}
\label{app:method}

\textbf{Sketched Epps--Pulley statistic.} For scalar samples $p_1,\dots,p_B$ (one random projection of the batch at one position), let $\hat\varphi(t) = \frac{1}{B}\sum_b e^{itp_b}$ be the empirical characteristic function and $\varphi(t) = e^{-t^2/2}$ the standard-normal one. The statistic is
\[
\mathrm{EP}(p_{1:B}) \;=\; B \int_{-3}^{3} \big|\hat\varphi(t) - \varphi(t)\big|^2\, \varphi(t)\, dt,
\]
with $\varphi(t)$ acting as the window (LeJEPA's default bandwidth) and the integral evaluated by a 17-knot trapezoid rule on $[0,3]$ using the integrand's evenness; the $B$-scaling removes the leading $1/B$ dependence of the statistic. $\mathrm{SIGReg}(h)$ at position $t$ averages $\mathrm{EP}$ over $m$ fresh unit directions and the loss is the mean over positions (Algorithm~\ref{alg:sigreg}); $m{=}1024$ in pretraining, and the A1 probe's per-layer \kv terms use $m{=}256$ (memory-bound; $m$ affects only the new term's sketch, and directions resample every step). The implementation is verified line-for-line against the LeWM reference implementation.

\textbf{Probe objectives.} A1 trains CE $+\,\lambda_z\cdot\mathrm{SIGReg}(h) +\lambda_{KV}\cdot\sum_\ell[\mathrm{SIGReg}(K_\ell)+\mathrm{SIGReg}(V_\ell)]$ with $\lambda_z{=}0.01$ (the checkpoint's own operating dose, unchanged), $\lambda_{KV}{=}0.01$, AdamW at lr $10^{-4}$, 500M tokens, bf16 autocast; the $\lambda_{KV}{=}0$ continuation control is the identical run with the \kv term removed. A2's gate metric is the next-position relative residual $\|X_{2:T} - P_\ell(X_{1:T-1})\|_F \,/\, \|X_{2:T}\|_F$ for $X_\ell \in \{K_\ell, V_\ell\}$, token-weighted and averaged over layers and $\{K,V\}$. A3 reconstructs $K_\ell/V_\ell$ from the final hidden state and is gated on the Frobenius error of the implied attention scores.

\textbf{Anisotropy sampling protocol.} Anisotropy$(\cdot)$ is the mean pairwise cosine over $10{,}000$ pairs drawn from a $4{,}096$-vector reservoir sampled over the held-out window under a fixed eval seed---the same protocol at the $h$ level and per-layer \kv level.

\textbf{Margin uncertainty (paired packed-batch bootstrap).} Group-free per-channel scales couple the eight packed sequences of an eval batch through the shared scale, so the resampling unit is the \emph{packed batch}, not the token or document. For every load-bearing 3-/4-bit configuration we replay the eval window recording per-batch (NLL sum, token count); the eval stream is deterministic, so batch $i$ is the same data for every model and configuration. Each of $10{,}000$ replicates draws one batch-index resample shared across all models and configurations (pairing), recomputes token-weighted NLL, and \emph{reselects each model's best configuration inside the replicate}---holding the observed winners fixed would ignore best-configuration selection noise. We report percentile intervals and bootstrap win frequencies under both relative-to-own-reference and absolute-NLL accounting (\texttt{scripts/kv\_quant\_bootstrap.py}).

\textbf{Channel-statistics protocol (\S\ref{sec:metric}).} Per layer and tensor, raw per-channel moments (orders 1--4, float64) stream over a 500k-token pass of the same held-out window: scale dispersion is max/median of per-channel standard deviations; excess kurtosis is the population central-moment ratio $m_4/m_2^2 - 3$ per channel; the token-norm concentration ratio is the median over batches of max/median per-token $L_2$ norm.

\section{Reproducibility notes}
\label{app:repro}

Pre-specification: bridge-probe gates fixed in the research plan (v12), committed 2026-04-25 in the repository history (a repository commit, not an externally timestamped registration); probe runs June--July 2026; the quantization study postdates the plan and is exploratory. Two software stacks: May-2026 cluster venv (dose--response pretraining) and a July rebuild (all paired pretraining results); all probes and sweeps ran on the local stack. Seed-42 reproduction across stacks shifted the two models $-1.78\%/{-1.90\%}$ ppl together; no cross-stack numbers are pooled anywhere. Evaluation determinism: fixed eval seed before every evaluation; train/eval streams disjoint by construction. Artifact caveat: the original A2/A3 run configs carry a vestigial \texttt{unfreeze:\,"trunk"} field from a shared CLI template---the trunk was frozen in code (gradients disabled at probe construction; pinned by tests), and probe configs generated after 2026-07-10 record the effective policy explicitly (\texttt{trunk\_policy:\,"frozen"}). Environment: frozen package listings for both stacks ship in the artifact bundle (\texttt{env/}); the dataset and tokenizer are referenced by name (upstream revision pins were not recorded at run time), and some audit configs record a dirty working tree without preserving the diff---exact bit-level re-execution is therefore bounded, and the artifact numbers, not re-execution, are the ground truth for every table.

\end{document}